OXFORD

# SpanSeq: similarity-based sequence data splitting method for improved development and assessment of deep learning projects

Alfred Ferrer Florensa [ID][1,*], Jose Juan Almagro Armenteros [ID][2], Henrik Nielsen [ID][3], Frank Møller Aarestrup [ID][1] and Philip Thomas Lanken Conradsen Clausen [ID][1]

[1]Research Group for Genomic Epidemiology, DTU National Food Institute, Technical University of Denmark, Anker Engelunds Vej 1, 2800 Kongens Lyngby, Denmark
[2]Informatics and Predictive Sciences Research, Bristol Myers Squibb Company, Calle Isaac Newton 4, 41092 Sevilla, Spain
[3]Section for Bioinformatics, Department of Health Technology, Technical University of Denmark, Anker Engelunds Vej 1, 2800 Kongens Lyngby, Denmark
*To whom correspondence should be addressed. Tel: +45 35 88 70 00; Email: alff@dtu.dk

## Abstract

The use of deep learning models in computational biology has increased massively in recent years, and it is expected to continue with the current advances in the fields such as Natural Language Processing. These models, although able to draw complex relations between input and target, are also inclined to learn noisy deviations from the pool of data used during their development. In order to assess their performance on unseen data (their capacity to *generalize*), it is common to split the available data randomly into development (train/validation) and test sets. This procedure, although standard, has been shown to produce dubious assessments of *generalization* due to the existing similarity between samples in the databases used. In this work, we present SpanSeq, a database partition method for machine learning that can scale to most biological sequences (genes, proteins and genomes) in order to avoid data leakage between sets. We also explore the effect of not restraining similarity between sets by reproducing the development of two state-of-the-art models on bioinformatics, not only confirming the consequences of randomly splitting databases on the model assessment, but expanding those repercussions to the model development. SpanSeq is available at https://github.com/genomicepidemiology/SpanSeq.

## Introduction

The value of deep learning models resides in their ability to predict accurately on previously unseen data. To achieve this *generalization* capacity, they rely on their considerable plasticity to learn underlying patterns in the data available, regardless of their complexity. Moreover, this potential allows them to *memorize* individual samples, perfectly fitting input with output of the training data, while decreasing the loss of the training set. The capacity to *memorize* by neural networks should not be underestimated, as they have even been able to fit random artificial data perfectly (1). On real cases, noisy data (2), out-of-distribution samples (3) and the size of the neural networks (2–5) seem to be key players in this issue. The repeated exposure to the same samples (3), which is a fundamental part of training, or the existence of duplicated samples in the dataset (6), are shown to help neural networks memorize.

*Memorization* tends to happen earlier during training than another problematic occurrence, *overfitting* (4). While *memorization* defines particular samples (mapping input to output), *overfitting* describes the fitting of the particularities of the training set, considering them as underlying patterns of the true distribution of data. These two phenomena should not be confused, as methods to control *overfitting*, like dropout or weight decay, show very limited effect on controlling the *memorization* (3). Moreover, the consequences on the useful-ness of the model differ. *Overfitting* should be displayed at the difference of the model's performance on training and test sets, which will be notably lower on the latter (7). When that happens, the value of the model is dubious, as its *generalization* capacity is not optimal. Nevertheless, the effect of *memorization* on the efficiency of a deep learning model is ambiguous (8). Being able to memorize out-of-distribution samples can be beneficial, particularly with certain data distributions (9,10), but it also provokes complications, as with privacy issues in language models (6,11), or by confusing *generalization* with *memorization*, overestimating the efficiency of a model (12).

A long-established procedure when developing a deep learning model is to split the available data in different sets, which are employed for different purposes: for fitting the model (training set), tuning its hyperparameters (training and development or validation set) and evaluating its performance (test set) (13). Thus, training and validation sets are used during the development of the model, using the latter to evaluate the model aiming to choose a set of hyperparameters that provides a better model, or to stop the training. Meanwhile, the test set will be used to evaluate the performance of the model (14), to have an estimation of how its performance will be once deployed.

The most common strategy for partitioning the data in different splits is to do it randomly (with respect to the input of the model), as an attempt to avoid any influence from the developer's previous knowledge of the dataset. This strategy is based on the assumption that the data is independent and identically distributed (i.i.d.) (15), which is not always the case. Duplicates and similarity among samples populate the current databases, and they spread among the train/validation/test sets when created by random splitting as data leakage. When that happens, similar points and duplicates appear in both training and test sets, inducing data leakage. In other words, the prediction on the test will be able to be solved by memorization. This can easily lead to an overestimation of the model's capacity to generalize, as it gets confused with its ability to memorize. This event has been found when developing and testing deep learning models on images (16), text (12,17) and code (18).

Data leakage through similarity is not restricted to those types of data. Biological sequences, such as genes, proteins or genomes, due to the evolutionary relationships that exist among them, and the redundancies in their databases (19), are a solid candidate to this type of data leakage. In fact, it is well known that similarity between biological sequences usually leads to similar phenotypes (20,21). This has been the principle used in alignment-based prediction methods (22), but also an object of concern as it causes data leakage when developing machine learning models or other algorithms (23). If above certain similarity threshold ($\tau$) sequences have the same phenotype, it is impossible to know if our model is predicting based on *memorization* or on *generalization* if it is presented with a sequence similar above τ to a sequence seen in the dataset. In those cases, alignment methods might be a better option than deep learning (24).

Some strategies have been used in later years to address the existence of similarity in datasets of biological sequences, mainly performing *data reduction*, or using a similarity-aware *data partitioning*. The first, *similarity data reduction*, clusters the data around representative sample points of the data itself, which are always distant above a certain threshold of similarity. Afterwards, any sequence that is not a representative is removed from the database, thus no similar sequences are contained in the database when using any type of data partitioning, such as random split. This method has several benefits, such as solving the issue of similarity inside sets (18). But it can also bring negative consequences, such as bias from the data points selected as representative, being left with an insufficient amount of data for training the model, or waste possible beneficial effects on training similarity can bring (as adversarial examples (25)). The second, *similarity-aware data partitioning*, groups similar sequences in clusters, which will be distributed among the partitions, avoiding similar samples to be in different partitions. Those two strategies are not exclusive, as the first can be used for removing duplicates, and the second for removing any data leakage from less similar sequences. For the research in this paper, we have chosen to evaluate the latter, as it is a feasible step no matter the quantity of data available.

Grouping similar sequences is a common step in both similarity-aware *reduction* and *partition* strategies. In the field of bioinformatics, identity from pairwise alignment methods (26,27) has been largely used as a measure of similarity. However, these methods have a complexity of $\mathcal{O}(l_1 l_2)$, where $l_1$, $l_2$ are the lengths of the two sequences aligned. With this complexity, calculating all of the distances of a large dataset (as required for deep learning) with pairwise alignment can eas-ily become infeasible, especially with long sequences. In order to address large datasets, most of the clustering strategies for biological sequences attempt to reduce the amount of alignments, such as CD-HIT (28), uCLUST (29), HHblits (30) and MMseqs (31), as per data reduction. While some of those methods are adequate for data reduction, due to using the representatives of each cluster for the similarity threshold, they are not suitable for data partition. A few methods have recently addressed this issue, such as attempting to create appropriate training and test sets of biological sequences (32), split molecular databases to reduce data leakage (DataSAIL (33)), or partition databases using graph-based approaches (GraphPart (34)). Although all of these methods split the data necessary for deep learning development at a reasonable time, as alignment-based approaches their usefulness is limited by sequence lengths.

In this work, we present SpanSeq, an alignment-free approach to similarity-aware data partitioning for deep learning models, able to handle large datasets of biological sequences of millions of units while still using an all-vs-all approach. Moreover, we apply this new method to the state-of-the-art protein subcellular location model DeepLoc (35,36) and to the method of RNA structure prediction DL-RNA (37), proving the importance of similarity-aware data splitting not only during model assessment, but through the whole model development process.

## Materials and methods

### SpanSeq method

The SpanSeq method is based on three steps:

(1) *Similarity calculation* among all the sequences of the dataset. It is performed with Mash (38) (amino acids and nucleotides) or KMA (39) (nucleotides).
(2) *Clustering* of similar sequences into clusters (DBSCAN)
(3) *Partitions creation* by distributing the clusters into $k$ partitions (Makespan) by minimizing the difference of the number of samples between partitions.

The software that forms SpanSeq has been implemented in C++, and organized using the workflow management system SnakeMake (40). The pipeline can be installed through the public Github repository https://github.com/genomicepidemiology/SpanSeq.git.

### Similarity calculation

As an alternative to pairwise alignment, SpanSeq takes advantage of $k$-mer comparisons, where sets of $k$-mers are compared and used as an unbiased estimate of the pairwise alignment (38). Although the comparison of $k$-mer sets is computationally efficient, when compared to pairwise alignment, it is still considered a computationally heavy process when the length of the sequences exceeds a few thousand bases (41). To overcome this computational burden, SpanSeq offers the possibility of sub-sampling $k$-mers using either MinHash (through MASH (38)), Minimizers (42) or $k$-mer prefixes (through KMA (39)). These techniques shrink the sizes of the individual $k$-mer sets, which makes the set comparisons faster while it requires less memory to compute. Currently, the distances between amino acid sequences can only be calculated by Mash distance (38), while distances between nucleotide sequences can be calculated by Mash [38], Cosine, Inverse K-Mer Coverage, Jaccard and Szymkiewicz-Simpson distance (Supplementary Material, Supplementary Equations S1, S2, S3, S4).

### Clustering

To avoid similar sequences between partitions, SpanSeq clusters similar sequences together using DBSCAN [43] (implemented in the software CCPhylo [44]). In SpanSeq, as the parameter *minPoint* (minimum number of close points to make a cluster) is set to 1, DBScan behaves as a single-linkage clustering method. The parameter $\epsilon$ (minimum distance between clusters) is used as the minimum distance between points to be located in different partitions. Due to the complexity of $\mathcal{O}(N^2)$ at the worst case [45], DBSCAN is suitable for most datasets up to a few hundred thousand data points. Nevertheless, to allow for massive data-clustering, a Hobohm 1 [19] clustering can optionally be applied with a runtime of $\mathcal{O}(MN)$ previously to the distance calculation step, where $N$ is the number of sequences and $M$ is the average sequence length. Notice that the complexity does not include the amount of clusters as with traditional cases ($\mathcal{O}(MNK)$, where $K$ is the amount of clusters), as constant lookups can be performed on k-tuples ($k$-mers) towards multiple samples at once when working with sequence (and text) data with a constant number of pass throughs of the data [39,46]. This step can either serve to collapse very close data points, or be applied as the main clustering approach, although the latter can introduce similarity between partitions.

### Partitions creation

In SpanSeq, the partitioning of clusters into separate cross-validation folds is treated as a *makespan* problem [47]. By default, it minimizes the difference of the amount of samples between partitions, while the optimization criteria can be changed at runtime to weigh cluster size differently (e.g. log- or squared-weights) and/or prioritize equal amounts of clusters between partitions with a secondary minimization of total size difference. Likewise, a minimization criterion based on class imbalance was added, to balance prediction labels between partitions, by minimizing the difference between all labels in each partition rather than just the total size. To minimize the makespan of cross-validation partitions, a tabu search algorithm [47] was implemented, which is initiated by a longest processing time solution (also known as decreasing best first solution). Through the tabu search, pairs of clusters are exchanged between partitions, where the optimal exchange between all partitions and clusters is accepted as the new best solution given that it is better than the previous best. This process is then repeated until no more exchanges that benefit the overall optimization criteria can be made. To limit the chances of getting stuck in a local minima, exchanges providing equally well suited solutions are accepted if they instead provide a heightened flexibility by minimizing the average amount of clusters. The search for an acceptable exchange requires $\mathcal{O}(P^2(C/P)^2) = \mathcal{O}(C^2)$ in each iteration (with $P$ being number of partitions, $C$ the number of clusters). However, by keeping the clusters sorted within each partition, the optimal exchange between any two partitions can be identified by a single pass-through of the clusters in the two partitions, giving a worst case runtime of $\mathcal{O}(P^2C/P) = \mathcal{O}(PC)$.

### Distance measures assessment

Three datasets were used for assessing the use of $k$-mer distances as similarity measure:

- Protein Sequences: The dataset from DeepLoc 2.0 [36] was chosen as the protein sequence case, since DeepLoc was going to be used as a test application to assess the importance of similarity-splitting. Due to the limitations of using the DeepLoc 1.0 model explained below, only 19,171 protein sequences from the dataset were used.
- Gene Sequences: The database of ResFinder [48] from (05/02/2023) was used as the genes dataset due to containing sequences of different length and uneven similarities.
- Genome Sequences: Multiple datasets were created from bacterial genomes from RefSeq [49] (06/04/2023), ranging from 10 to $10^5$. All of them consist of *complete genomes* exclusively (without their plasmids), with the exception of the largest, which had to also include genomes split in contigs.

In order to assess the use of the different $k$-mer distances, their correlation with identity was evaluated. Identity was reported by global pairwise alignment (performed with ggsearch36, part of the package FASTA36 (version: 36.3.8i) [50]). As global alignment was not feasible for the dataset of genomes, the clusters provided by SpanSeq were compared with the taxonomy of the samples.

### Performance evaluation

The performance evaluation was done on the genome dataset described above, in order to evaluate the method not only in large datasets (from $10^1$ to $10^5$), but also on sequences with lengths large enough to be prohibitive for most alignment methods. To run SpanSeq on that dataset, a machine with an AMD EPYC 7352 24-Core Processor (96 CPUs), and the OS system Ubuntu 22.04.3 LTS was used. Each CPU has 2100-1500 MHz speed, 2.0 TB of RAM Memory and 8.0 GB of RAM Swap.

### Similarity-based partition effect on deep learning models development

To evaluate the effect of similarity-based partitioning, the model DeepLoc [35] and the demonstrative model of [37], DL-RNA, were used. While the first one was developed from beginning to end (hyperparameter selection, training, model assessment), the second one was only trained and model assessed. In both cases, we followed the same procedure for splitting the model's databases, in order to uncover the effect of data similarity. We used a static threshold of similarity ($\tau$) as the point where data leakage would occur for each of the models based on the bibliography found, as explained below. Investigating the precision of those thresholds is outside the scope of this research, as previous research has shown it is problem-specific to what extent sequence similarity correlates to phenotype similarity [24]. The databases for each case were split following a two-step procedure to prove the usefulness of data similarity limitation methods (as SpanSeq), producing the following sets of data:

- **Hold out test set**: The dataset was originally partitioned in six subsets using SpanSeq (Mash distance [38] for DeepLoc, cosine distance for DL-RNA), to restrict the similarity between sequences in different sets to a maximum of τ. From that split, one set was randomly selected to be the hold out test set, while the other five were merged to be the development set.

- **Development set**: This set (5/6 of the data) was partitioned into five other sets: one for testing and four for training and validation set. However, four different partitioning methods were used, each of which with different approaches towards similarity between splits.

  - *SpanSeq (Mash/cosine) split:* The development set was split using SpanSeq, with the mash and cosine distances. Thus, there is no pair of sequences more than τ similar between two different sets of data.
  - *SpanSeq (pairwise alignment) split:* The development set is split using SpanSeq, with global pairwise alignment (with ggsearch36 (50)) for calculating the distance among sequences. Thus, there is no pair of sequences more than τ similar between two different sets of data.
  - *Random split:* The development set is split randomly, so there is no control of where similar sequences are placed.
  - *Increased similarity split:* The development set is split with similar sequences being spread in different partitions, so that the similarity is increased between them. This was done by inverting the distance measure between sequences, so that closely related sequences were treated as distantly related sequences and vice versa, thus spreading the closely related sequences into different partitions.

The hyperparameter selection (only on DeepLoc) was performed using a 4-fold cross-validation scheme, applying a Bayesian hyperparameter optimization (SigOpt (51)). Each of the different dataset partitions had the same parameters for hyperparameter selection (amount of combination tried, hyperparameters' ranges explored). The hyperparameter selection on DL-RNA (done by the authors) was not done following a similarity-aware method.

For both methods (DeepLoc and DL-RNA), each model was trained four times, with each time three partitions being used for training while one being used as a validation set. The training lasted for 800 epochs, and the model iteration with best performance on the validation set was selected as the trained model. This trained model is the one used for model assessment on the test and hold out test set, in order to assess the effect of similarity on classic methods for *overfitting* such as early-stopping.

As both methods are unbalanced classification models (52), the Matthews correlation coefficient (MCC) (53) (Supplementary Material, Supplementary Formula S5) and its generalization to multiclass problems, known as the Gorodkin measure (54) (Supplementary Material, Supplementary Formula S6), were used as performance measures for DL-RNA and DeepLoc, respectively.

The hyperparameter optimization and training of DeepLoc were performed on one node of 2 T V100 16 GB GPUs, while the training of DL-RNA was performed on one node of 2 T A100 PCIE 40 GB GPUs.

### DeepLoc

The model DeepLoc (35) is a deep neural network that predicts the subcellular location of eukaryotic proteins, using a combination of convolutional, LSTM, and attention layers. Although at the time of the SpanSeq research DeepLoc 2.0 (36) had been released, it was decided to evaluate on the first version, as the newest uses a pre-trained model. We believed that not having full control of which data the model has seen would make it difficult to evaluate the effect similarity between splits.

As DeepLoc works on amino acid sequences, a fixed threshold (τ) of 0.3 similarity was chosen, below which two sequences would not be considered close enough to be prone to data leakage. This number was chosen based on previous studies of the relationship between amino acid sequence and protein subcellular location (55,56) and its usage in deep learning models (35).

The dataset of DeepLoc 2.0 (36) was used for developing DeepLoc 1.0 (35), as it was newer and larger (36). Proteins with multiple locations, as well as the ones longer than 1,000 amino acids, were dropped due to DeepLoc 1.0's inability to classify these, unlike DeepLoc 2.0, leaving a dataset of 19,171 sequences.

### DL-RNA

The model DL-RNA (37) is a convolutional neural network that aims to predict the secondary structure of an RNA sequence. The sequence of nucleotides is one-hot encoded into matrices that the model uses to predict for each position if that nucleotide is paired (binary classification).

As the authors of DL-RNA consider that using a small database (ArchiveII (57), 2975 RNA structures) could have caused the poor performance of their model, we trained the model using the database RNAstralign (58) (37149 RNA structures). The database was homology reduced with a threshold of 0.98, and RNA sequences longer than 1048 nucleotides were dropped due to memory limitations on our GPUs.

The similarity threshold (τ) for RNA structure prediction is currently a matter of debate. Other studies (59–61) have used a strategy of data reduction with a τ of 0.8. A threshold of 0.7 for similarity-aware data partition has also been used (60). Nevertheless, the authors of DL-RNA (37), have found those measures insufficient, as RNA sequences from the same family should not appear in different partitions (which, in terms of identity, is below 0.6). However, due to the limited diversity in the database used, SpanSeq is unable to partition the data with a lower τ than 0.7, which is the value we ended using. We acknowledge that this value might not be sufficient, so some data leakage might have occurred.

## Results

### SpanSeq features

#### Distance measures and identity correlation

The correlation between SpanSeq's $k$-mer distance measures and identity from global alignment can be seen in Figure 1 and Supplementary Material, Supplementary Figure S3. Most distances display a high correlation with identity; except Jaccard distance (Supplementary Material, Supplementary Figure S3). Further exploration of the relation between the difference between $k$-mer distances and sequence lengths was performed (Supplementary Data, Supplementary Figure S4, as well as the effects of $k$-mer and minimizer size; although an in-depth study of them is outside the scope of this research.

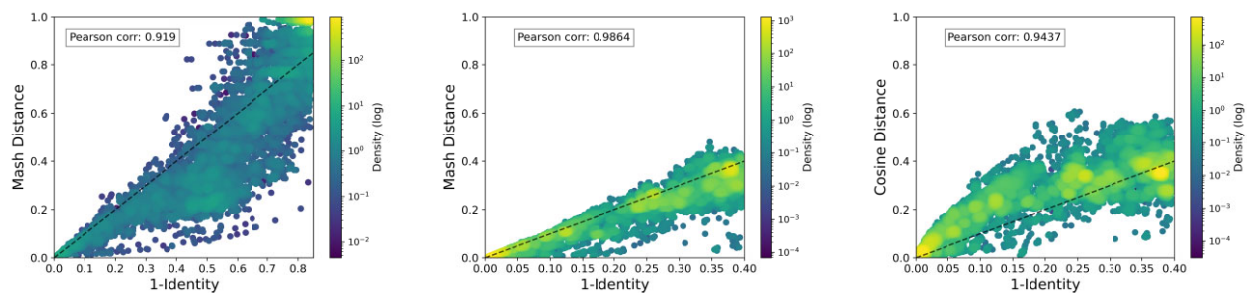

**A** Mash distance against global identity on protein dataset, using *k*-mer size 6 and scheme size of 1048.

**B** Mash distance against global identity on gene dataset, using *k*-mer size 8 and scheme size of 1048.

**C** Cosine distance against global identity on gene dataset, using *k*-mer size 8 and minimizer of 5.

**Figure 1.** Correlation between two out of the five available distances on SpanSeq, on the protein and gene datasets. The rest of the distances (Inverse K-Mer Coverage, Jaccard and Szymkiewicz-Simpson) show a similar behavior, as seen in the Supplementary Material, Supplementary Figure S3. The visualization is limited to the identity ranges 100% - 15% for amino acids, and 100–60% for nucleotides, due to the artifacts that appear when aligning distant sequences (62) (Supplementary Data, Supplementary Figures S1 and S2).

### Distance clustering and taxonomic comparison

The clustering results of SpanSeq on the $10^4$ genome dataset show differences between the different distance measures regarding the taxonomic relations among the genomes clustered (Figure 2). The Cosine, Szymkiewicz–Simpson and *K*-mer inverse coverage distances show similar distributions for the five NCBI taxonomic ranks. While the majority of the intra-species distances are between 0.2 and 0.4, the distances in the other taxonomic ranks are collapsed at the maximum value possible. The Jaccard distance shows a distribution consonant with its correlation with identity (Supplementary Material, Supplementary Figure S3F). With Mash distance, the distributions on the taxonomic ranks differ considerably from the other distances. Most of the intra-species distances are below 0.05, while the other ranks most show values ranging from 0 to 1.

### Performance evaluation

The results show that SpanSeq was able to partition datasets in a sensible time relative to the size of the datasets (Supplementary Data, Supplementary Figure S5) using an all-vs-all strategy and different strategies depending on the size of the dataset (minimizers/prefixes).

### Similarity-aware partitioning on DeepLoc and DL-RNA

#### Hyperparameter selection on DeepLoc

The hyperparameter configuration from each of the dataset partitions modes differs, although SigOpt (51) was run with the same range of hyperparameters in each case (Supplementary Data, Supplementary Table S1). Besides, the 4-fold cross-validation performance (Gorodkin score) seemed directly correlated to the expected amount of similarity between sets (Supplementary Data, Supplementary Table S1).

#### Training process

The training curves of DeepLoc (Figure 3) and DL-RNA (Supplementary Material, Supplementary Figure S6) differ among partition strategies. When training DeepLoc, the training using the increased similarity and random partition datasets fits rapidly the training data almost perfectly (Figure 3A, B), followed by a deceleration of its improvement. The performance on the validation set follows a similar course, but 0.2 points below. In the case of the datasets split with SpanSeq, the training curves increase sharply during the first 30 epochs, after which they keep incrementing slowly from there without reaching 0.9 during the 800 epochs trained (Figures 3C, D). The curve of the performance on the validation set, though, follows the training curve until it plateaus and starts decreasing after the 100th epoch, unlike the trainings done without SpanSeq, which plateau or even increase (Figures 3A, B). There are also differences on which epochs the performance of the validation set is higher. Those epochs appear certainly much earlier during the training of SpanSeq split datasets (vertical lines in Figure 3; Supplementary Material, Supplementary Figure S8).

Mostly the same patterns are observed also when training DL-RNA (Supplementary Material, Supplementary Figure S6) with the different data partition strategies. However, on the datasets split with SpanSeq (and to a lesser extent on the random split), some of the performances on the validation sets vary noticeably, with some of them reaching values close to the ones of the increased similarity and random partition. Besides, although the epochs with the highest performance on the validation set still appear earlier during the training process, the training curves between data partition strategies are very similar.

#### Model evaluation

When evaluating the performance of DeepLoc (Figure 4) and DL-RNA (Supplementary Material, Supplementary Figure S7), we observe differences correlated to those seen in the training curves. Furthermore, the performances on the test and test holdout sets show differences between them depending on the data partition strategy. Using SpanSeq, the performances on the test and test holdout are almost equal, as on the validation set. When using the increased similarity or random partitions, the validation and test sets show a notably higher performance than on the test holdout, which are slightly smaller than on the models developed with SpanSeq.

## Discussion

### Sequence similarity calculation using distance measures

SpanSeq is able to use a clustering strategy of all-vs-all on large datasets within a reasonable amount of computational re-sources (Supplementary Material, Supplementary Figure S5). Using *k*-mer comparison instead of alignment related measures allows for this. The results of the correlation (Figure 1, and Supplementary Material, Supplementary Figure S3), indicate that most *k*-mer distances gather information similar to that retrieved by identity. In fact, the correlation could be even higher than as seen in our results, as the difference between both measures increases as identity decreases and is more prone to artifacts (Supplementary Material, Supplementary Figures S1 and S2). *K*-mer distances are also unaffected by recombination events (unlike global alignment), which could explain some of the points where *k*-mer distances show a much smaller value than the inverse of the identity.

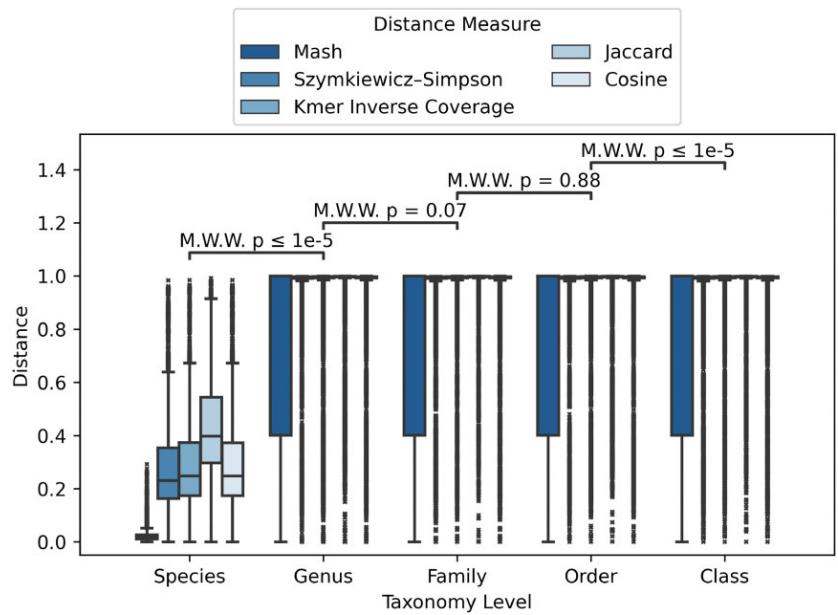

**Figure 2.** Boxplots of distances between genome sequences inside the same taxonomic level. Mash distance was calculated with *k*-mer size 19 and scheme size 4096, while the rest were calculated using *k*-mer size 18 and minimizer size 16. The results of the Mann–Whitney–Wilcoxon *U* test (MWW) [63] on the distance distributions show unequal results depending on the ranks compared.

The choice of hyperparameters of SpanSeq (*k*-mer, minimizer, etc.) can affect the correlation between distance and identity. Moreover, there are clear relationships between their optimal values and the length of the sequences to be clustered (Supplementary Material, Supplementary Figure S4). However, a study of strategies for choosing these hyperparameters is outside of the scope of this study.

The use of *k*-mer distances as a measure of similarity makes it possible to standardize methods for avoiding data leakage when working with long sequences such as genomes. Previously limited to using the taxonomic nomenclature, partitions made with SpanSeq should be more consistent, while they show clear correlations (Figure 2). Besides Jaccard and Mash, the *k*-mer distance measures show a very similar behavior. Although they are unable to distinguish taxonomic ranks above species, almost all the intra-species distances are <0.7, making it a good indicator for a species-aware data partitioning. In fact, inspecting the clusters formed on the genome dataset using SpanSeq, deficiencies on the taxonomic annotations arise: the clusters with multiple species are, in fact, the same bacteria species, or species difficult to differentiate [64] (Supplementary Material, Supplementary Table S2).

While Jaccard distance overestimates the distance between sequences (of any length), Mash shows a behavior not observed with shorter sequences, underestimating the distance between very dissimilar long sequences. This is a result of a logarithmic error profile produced by the Jaccard estimate calculated through Mash, giving a small error-margin of similar sequences that grows as the sequences become more dissimilar.

## Effect of data similarity on deep learning models

The clustering (and the following makespan step) performed by SpanSeq does not search for a representative clustering of the dataset, as other algorithms [28], but aims to build independent partitions and avoid data leakage between them. The outcome of this dataset partition strategy has been tested by training DeepLoc and DL-RNA with four different approaches that show clear differences on the test and validation sets performances. Seeing such variation is remarkable, as the DeepLoc2.0 dataset does not contain a large number of similar sequences (Figure 1A), and the RNAStralign database has been homology reduced.

Notice that both models are single sequence predictors with previous research on similarity thresholds for data leakage. This makes them suitable candidates to benefit from similarity-aware partitions. The use of SpanSeq for input data that are built as a conglomerate of different sequences, as sequence profiles or proteins with their ligands, has not been explored in this research.

## Model assessment

The purpose of using a separate test set when evaluating a model is to assess the generalization capacity of our model, as its data should be different to those seen during training. In this case, for each partition we have a test set (which is a common procedure in the development of a model) and a hold out

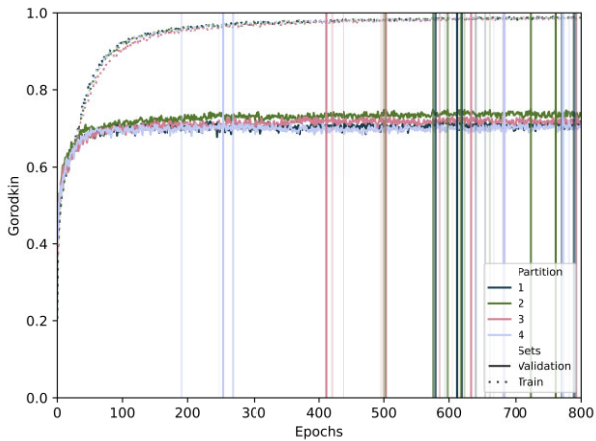

**A** Training process with increased similarity split dataset. The epochs 612, 762, 411, 253 have the best validation set Gorodkin measure (0.72, 0.75, 0.74, 0.72; respectively)

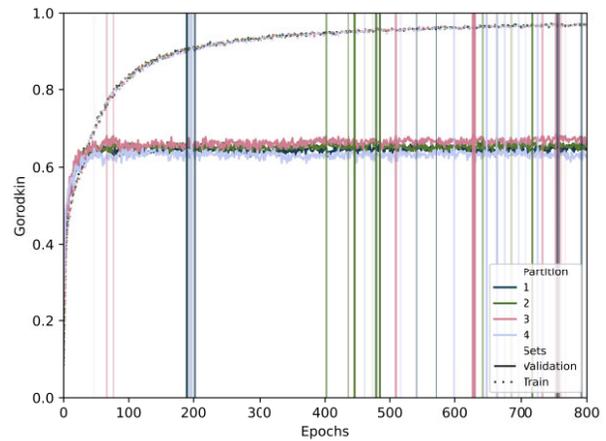

**B** Training process with randomly split dataset. The epochs 780, 276, 533, 193 have the best validation set Gororodkin measure (0.65, 0.68, 0.69, 0.64; respectively)

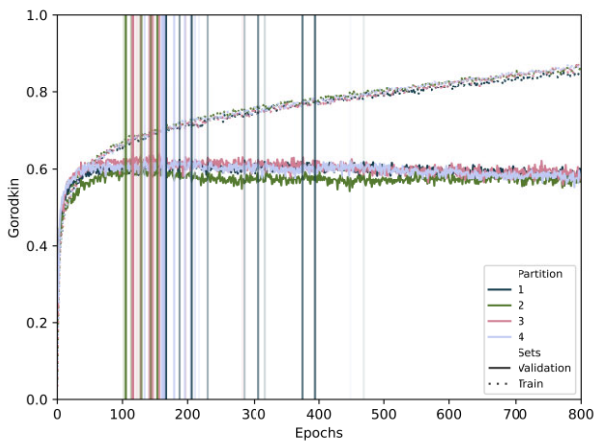

**C** Training process with SpanSeq (Mash) split dataset. The epochs 165, 105, 116, 162 have the best validation set Gorodkin measure (0.62, 0.60, 0.64, 0.62; respectively)

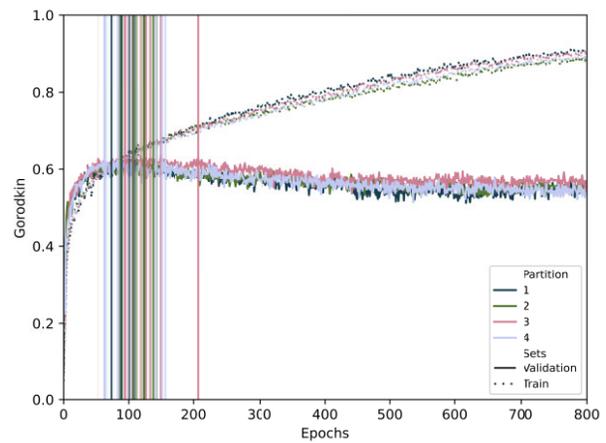

**D** Training process with SpanSeq (GGSearch) split dataset. The epochs 123, 67, 186, 105 have the best validation set Gorodkin measure (0.64, 0.62, 0.62, 0.62; respectively) test set to evaluate that test set. In doing so, it is found that the test sets created without a similarity-aware method (increased homology and random) overestimate the model's generalization capacity (Figure 4). This behavior coincides with previous studies done on similarity between the training and the test set (12), questioning if a test set created with random split is a truthful method to report the capacity to generalize of a model (especially when working with sequential data). In fact, it is reasonable to claim that the difference between the performance values on the test and hold out test sets of the random and increased homology partitions comes from similar sequences between training and test sets (that the hold out test set is free from). Thus, when evaluating the model, some of the model's generalization ability that is claimed is, indeed, its memorization capacity.

**Figure 3.** Training curves for the different datasets. The dashed and continuous lines show the Gorodkin measures of the train and validation sets respectively. The vertical lines show the ten best validation Gorodkin values for each validation set (the transparency of these lines is inversely related to the validation Gorodkin value). The best iteration (vertical line without transparency) is the iteration used to evaluate the performance on the test set.

There is a lower performance on the holdout test set when not using similarity-aware partitions (at least on DeepLoc, where more research has been done on similarity threshold), which could be explained by the overfitting control methods being compromised. However, the difference between the per-formances among the partition methods are not sufficient to make that statement.

## Model development

On similarity non-aware partitions, the performance on some of the validation set also shows higher values compared to the hold out set, as in the model assessment. Methods depending on a validation set are likely going to be affected by the presence of similar sequences with the training set. A clear example is the method to avoid *overfitting*, early-stopping. When using similarity-aware partitions in DeepLoc, the Gorodkin measure of the validation set follows the one of the training set, until it plateaus and slightly diminishes as the epochs increase. This common dynamic shows the neural network starting to *overfit* during training, as fitting the particularities and noise from the training set, in detriment to fitting the real distribution of the data. Early-stopping would act then, impeding the training to continue. However, on the other two dataset partitions, the validation performance does not decrease (even increases slightly), as some of the particularities on the training data are shared in the validation set. Thus, making early-stopping, and any other method based around a validation set, dysfunctional.

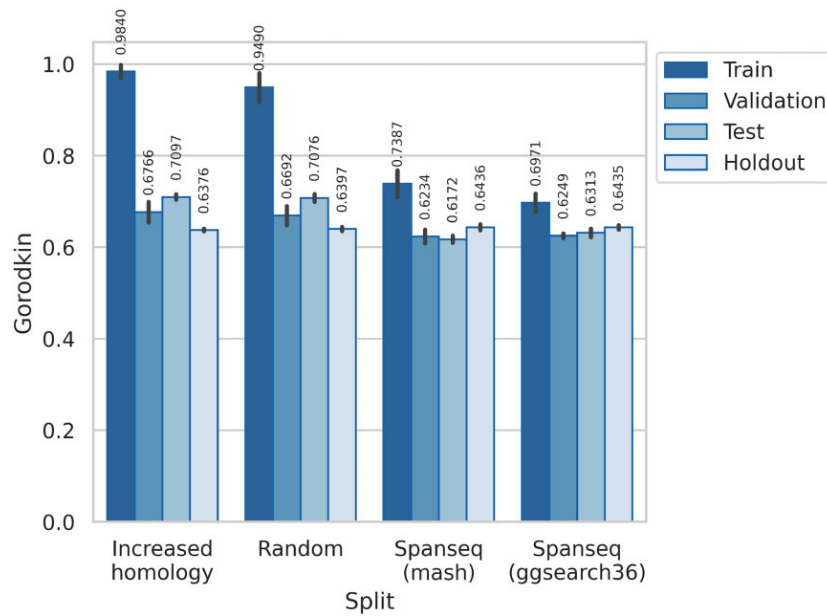

**Figure 4.** Gorodkin measure on the different sets depending on the split method, with the value on top of each column. The vertical line on top of each bar displays the standard deviation. Although training and validation performances are calculated using a *k*-fold scheme, the test set and hold out set are not. This might explain some of the short differences in performance between them, due to sets with unequal difficulty for the model.

*K*-fold cross validation is one of them. The training curves of the models with hyperparameter selection done with partitions that are not similarity-aware (Figure 3A, C, and Supplementary Material, Supplementary Figure S6) show very similar dynamics. In fact, they suggest that configurations where the fitting of the training data are endorsed by the hyperparameter selection, as the performance on the validation set will benefit from it due to *memorization*. This dynamic is not seen on the DeepLoc models with similarity-aware partition before hyperparameter selection.

It is noticeable that in some cases (increased homology and random split in DeepLoc, for example) the performance in the validation set is not uniform (shown with a large s.d. in Figure 4). In fact, some of those sets perform worse than the test set. This can be explained by an unequal difficulty for the model among validation sets, but also by an unequal distribution of the similar sequences. The same pattern is found in random and SpanSeq splits in DL-RNA (Supplementary Material, Supplementary Figure S7), of which we suspect there is some similarity between sets due to a similarity threshold (τ) not restrictive enough.

The effects of similarity between sequences are not limited to the model's performance on the different sets, as the time necessary for developing the deep learning model (hyperparameter selection and training) can be reduced drastically. As seen in Figure 3 and Supplementary Material, Supplementary Figure S6, the epochs that can generalize better appear dramatically earlier when using similarity-based splitting (SpanSeq). This fact can be applied not only at the training process (with early stopping or reducing the amount of epochs that the model is trained for), but also for hyperparameter selection (amount of epochs necessary for assessing the parameter configuration). Considering the power consumption for training deep learning models (65), the use of SpanSeq can help making their development much more ecologically sustainable and economically affordable for smaller teams.

## Conclusion

In this work, we present SpanSeq, a dataset partition method specially created for the development of deep learning models. Unlike other clustering methods for biological sequences (28), SpanSeq uses a clustering method based on *k*-mer distances and all-vs-all clustering to assess data leakage from similarity during data partition. This provides balanced partitions with respect to their size, as well as an option to minimize class imbalance. Due to its efficient distance calculation, the method is able to handle large datasets of biological sequences, no matter their length (proteins, genes and genomes).

Furthermore, we prove the need to use similarity-aware partitions such as SpanSeq over random splits. First, to prevent reporting overconfident generalization capacity of the model, as the performance on the test set will be partially due to its memorization capacity. Second, to protect methods based on the performance on validation sets from data leakage, so they benefit models with *generalization* capacity. And last, to largely reduce the resources necessary for developing a deep learning model, without compromising the generalization capacity of the model.

## Data availability

The software SpanSeq is available at the Github repository https://github.com/genomicepidemiology/SpanSeq.git and 10.1093/nargab/lqae106. SpanSeq uses the softwares CCPhylo [9], KMA [10] and Mash [41]. The software for DeepLoc 1.0 [2] is available at https://github.com/JJAlmagro/subcellular_localization and the database of DeepLoc 2.0 [58] is available at https://services.healthtech.dtu.dk/services/DeepLoc-2.0/. The software for DL-RNA [53] is available at

https://github.com/marcellszi/dl-rna/, while the database RNAStralign is available at https://rna.urmc.rochester.edu/.

## Supplementary data

Supplementary Data are available at NARGAB Online.

## Funding

European Union's Horizon 2020 research and innovation program under VEO grant [874735].

## Conflict of interest statement

Jose Juan Almagro Armenteros is an employee of Bristol Myers Squibb Company at the time of the publication; however, that did not influence the research in any way.

## References

1. Zhang,C., Bengio,S., Hardt,M., Recht,B. and Vinyals,O. (2021) Understanding deep learning (still) requires rethinking generalization. *Commun. ACM*, **64**, 107–115.
2. Arpit,D., Jastrzębski,S., Ballas,N., Krueger,D., Bengio,E., Kanwal,M.S., Maharaj,T., Fischer,A., Courville,A., Bengio,Y., *et al.* (2017) A closer look at memorization in deep networks. In: *International Conference on Machine Learning*. PMLR. Vol. 70, pp. 233–242.
3. Carlini,N., Liu,C., Erlingsson,Ú., Kos,J. and Song,D. (2019) The secret sharer: Evaluating and testing unintended memorization in neural networks. In: *28th USENIX Security Symposium (USENIX Security 19)*. pp. 267–284.
4. Tirumala,K., Markosyan,A., Zettlemoyer,L. and Aghajanyan,A. (2022) Memorization without overfitting: Analyzing the training dynamics of large language models. *Adv. Neur. Inf. Proc. Syst.*, **35**, 38274–38290.
5. Zhang,C., Bengio,S., Hardt,M., Mozer,M.C. and Singer,Y. (2020) Identity crisis: memorization and generalization under extreme overparameterization. In: *International Conference on Learning Representations*.
6. Carlini,N., Ippolito,D., Jagielski,M., Lee,K., Tramer,F. and Zhang,C. (2022) Quantifying memorization across neural language models. In: *The Eleventh International Conference on Learning Representations*.
7. Tetko,I.V., Livingstone,D.J. and Luik,A.I. (1995) Neural network studies. 1. Comparison of overfitting and overtraining. *J. Chem. Inf. Comput. Sci.*, **35**, 826–833.
8. Chatterjee,S. (2018) Learning and memorization. In: *International conference on machine learning*. PMLR, pp. 755–763.
9. Feldman,V. and Zhang,C. (2020) What neural networks memorize and why: Discovering the long tail via influence estimation. *Adv. Neur. Inf. Proc. Syst.*, **33**, 2881–2891.
10. Feldman,V. (2020) Does learning require memorization? a short tale about a long tail. In: *Proceedings of the 52nd Annual ACM SIGACT Symposium on Theory of Computing*. pp. 954–959.
11. Lee,K., Ippolito,D., Nystrom,A., Zhang,C., Eck,D., Callison-Burch,C. and Carlini,N. (2022) Deduplicating training data makes language models better. In: *Proceedings of the 60th Annual Meeting of the Association for Computational Linguistics*. Association for Computational Linguistics. pp. 8424–8445.
12. Elangovan,A., He,J. and Verspoor,K. (2021) Memorization vs. generalization: quantifying data leakage in NLP performance evaluation. In: *Proceedings of the 60th Annual Meeting of the Association for Computational Linguistics*. Association for Computational Linguistics. pp. 8424–8445.
13. Hastie,T., Friedman,J.H. and Tibshirani,R. (2009) In: *The Elements of Statistical Learning: Data Mining, Inference, and Prediction*. Vol. 2, Springer.
14. Westerhuis,J.A., Hoefsloot,H.C., Smit,S., Vis,D.J., Smilde,A.K., van Velzen,E.J., van Duijnhoven,J.P. and van Dorsten,F.A. (2008) Assessment of PLSDA cross validation. *Metabolomics*, **4**, 81–89.
15. Vapnik,V.N. (1995) In: *The Nature of Statistical Learning Theory*. Springer.
16. Tampu,I.E., Eklund,A. and Haj-Hosseini,N. (2022) Inflation of test accuracy due to data leakage in deep learning-based classification of OCT images. *Sci. Data*, **9**, 580.
17. Søgaard,A., Ebert,S., Bastings,J. and Filippova,K. (2020) We need to talk about random splits. In: *Proceedings of the 16th Conference of the European Chapter of the Association for Computational Linguistics*. Association for Computational Linguistics, pp. 1823–1832.
18. Allamanis,M. (2019) The adverse effects of code duplication in machine learning models of code. In: *Proceedings of the 2019 ACM SIGPLAN International Symposium on New Ideas, New Paradigms, and Reflections on Programming and Software*. pp. 143–153.
19. Hobohm,U., Scharf,M., Schneider,R. and Sander,C. (1992) Selection of representative protein data sets. *Protein Sci.*, **1**, 409–417.
20. Lund,O., Frimand,K., Gorodkin,J., Bohr,H., Bohr,J., Hansen,J. and Brunak,S. (1997) Protein distance constraints predicted by neural networks and probability density functions.. *Protein Eng.*, **10**, 1241–1248.
21. Sander,C. and Schneider,R. (1991) Database of homology-derived protein structures and the structural meaning of sequence alignment. *Proteins Struct. Funct. Bioinform.*, **9**, 56–68.
22. Pearson,W.R. (2013) An introduction to sequence similarity ('homology') searching. *Curr. Protoc. Bioinform.*, **Chapter 3**, 3.1.1–3.1.8.
23. Nielsen,H., Engelbrecht,J., von Heijne,G. and Brunak,S. (1996) Defining a similarity threshold for a functional protein sequence pattern: the signal peptide cleavage site. *Proteins Struct. Funct. Bioinform.*, **24**, 165–177.
24. Li,Y., Sackett,P.W., Nielsen,M. and Barra,C. (2023) NetAllergen, a random forest model integrating MHC-II presentation propensity for improved allergenicity prediction. *Bioinform. Adv.*, **3**, vbad151.
25. Goodfellow,I.J., Shlens,J. and Szegedy,C. (2015) Explaining and harnessing adversarial examples. In: *International Conference on Learning Representations*.
26. Needleman,S.B. and Wunsch,C.D. (1970) A general method applicable to the search for similarities in the amino acid sequence of two proteins. *J. Mol. Biol.*, **48**, 443–453.
27. Gotoh,O. (1982) An improved algorithm for matching biological sequences. *J. Mol. Biol.*, **162**, 705–708.
28. Fu,L., Niu,B., Zhu,Z., Wu,S. and Li,W. (2012) CD-HIT: accelerated for clustering the next-generation sequencing data. *Bioinformatics*, **28**, 3150–3152.
29. Prasad,D.V., Madhusudanan,S. and Jaganathan,S. (2015) uCLUST – a new algorithm for clustering unstructured data. *ARPN J. Eng. Appl. Sci.*, **10**, 2108–2117.
30. Remmert,M., Biegert,A., Hauser,A. and Söding,J. (2012) HHblits: lightning-fast iterative protein sequence searching by HMM-HMM alignment. *Nat. Methods*, **9**, 173–175.
31. Hauser,M., Steinegger,M. and Söding,J. (2016) MMseqs software suite for fast and deep clustering and searching of large protein sequence sets. *Bioinformatics*, **32**, 1323–1330.
32. Petti,S. and Eddy,S.R. (2022) Constructing benchmark test sets for biological sequence analysis using independent set algorithms. *PLoS Comput. Biol.*, **18**, e1009492.
33. Joeres,R., Blumenthal,D.B. and Kalinina,O.V. (2023) DataSAIL: Data Splitting Against Information Leakage. bioRxiv doi: https://doi.org/10.1101/2023.11.15.566305, 17 November 2023, preprint: not peer reviewed.

34. Teufel,F., Gíslason,M.H., Almagro Armenteros,J.J., Johansen,A.R., Winther,O. and Nielsen,H. (2023) GraphPart: Homology partitioning for biological sequence analysis. *NAR Genom. Bioinform.*, **5**, lqad088.

35. Almagro Armenteros,J.J., Sønderby,C.K., Sønderby,S.K., Nielsen,H. and Winther,O. (2017) DeepLoc: prediction of protein subcellular localization using deep learning. *Bioinformatics*, **33**, 3387–3395.

36. Thumuluri,V., Almagro Armenteros,J.J., Johansen,A.R., Nielsen,H. and Winther,O. (2022) DeepLoc 2.0: multi-label subcellular localization prediction using protein language models. *Nucleic Acids Res.*, **50**, W228–W234.

37. Szikszai,M., Wise,M., Datta,A., Ward,M. and Mathews,D.H. (2022) Deep learning models for RNA secondary structure prediction (probably) do not generalize across families. *Bioinformatics*, **38**, 3892–3899.

38. Ondov,B.D., Treangen,T.J., Melsted,P., Mallonee,A.B., Bergman,N.H., Koren,S. and Phillippy,A.M. (2016) Mash: fast genome and metagenome distance estimation using MinHash. *Genome Biol.*, **17**, 132.

39. Clausen,P.T., Aarestrup,F.M. and Lund,O. (2018) Rapid and precise alignment of raw reads against redundant databases with KMA. *BMC Bioinformatics*, **19**, 307.

40. Mölder,F., Jablonski,K.P., Letcher,B., Hall,M.B., Tomkins-Tinch,C.H., Sochat,V., Forster,J., Lee,S., Twardziok,S.O., Kanitz,A., *et al.* (2021) Sustainable data analysis with Snakemake. *F1000Research*, **10**, 33.

41. Clausen,P.T., Zankari,E., Aarestrup,F.M. and Lund,O. (2016) Benchmarking of methods for identification of antimicrobial resistance genes in bacterial whole genome data. *J. Antimicrob. Chemoth.*, **71**, 2484–2488.

42. Li,H. (2016) Minimap and miniasm: fast mapping and de novo assembly for noisy long sequences. *Bioinformatics*, **32**, 2103–2110.

43. Ester,M., Kriegel,H.-P., Sander,J. and Xu,X. (1996) A density-based algorithm for discovering clusters in large spatial databases with noise. In: *Proceedings of the Second International Conference on Knowledge Discovery and Data Mining*. Vol. **96**. AAAI Press, pp. 226–231.

44. Clausen,P.T. (2023) Scaling neighbor joining to one million taxa with dynamic and heuristic neighbor joining. *Bioinformatics*, **39**, btac774.

45. Gan,J. and Tao,Y. (2015) DBSCAN revisited: Mis-claim, un-fixability, and approximation. In: *Proceedings of the 2015 ACM SIGMOD International Conference on Management of Data*. pp. 519–530.

46. Wilbur,W.J. and Lipman,D.J. (1983) Rapid similarity searches of nucleic acid and Protein Data Banks. *Proc. Natl. Acad. Sci. U.S.A.*, **80**, 726–730.

47. Hübscher,R. and Glover,F. (1994) Applying tabu search with influential diversification to multiprocessor scheduling. *Comput. Oper. Res.*, **21**, 877–884.

48. Bortolaia,V., Kaas,R.S., Ruppe,E., Roberts,M.C., Schwarz,S., Cattoir,V., Philippon,A., Allesoe,R.L., Rebelo,A.R., Florensa,A.F.,

*et al.* (2020) ResFinder 4.0 for predictions of phenotypes from genotypes. *J. Antimicrob. Chemoth.*, **75**, 3491–3500.

49. O'Leary,N.A., Wright,M.W., Brister,J.R., Ciufo,S., Haddad,D., McVeigh,R., Rajput,B., Robbertse,B., Smith-White,B., Ako-Adjei,D., *et al.* (2016) Reference sequence (RefSeq) database at NCBI: current status, taxonomic expansion, and functional annotation. *Nucleic Acids Res.*, **44**, D733–D745.

50. Pearson,W.R. (2016) Finding protein and nucleotide similarities with FASTA. *Curr. Protoc. Bioinform.*, **53**, 3–9.

51. Hayes,P., Anderson,D., Cheng,B., Spriggs,T.J., Johnson,A. and McCourt,M. (2019). SigOpt documentation. Technical Report SO-12/14 – Revision 1.07, SigOpt, Inc.

52. Chicco,D., Tötsch,N. and Jurman,G. (2021) The Matthews correlation coefficient (MCC) is more reliable than balanced accuracy, bookmaker informedness, and markedness in two-class confusion matrix evaluation. *BioData Min.*, **14**, 13.

53. Matthews,B.W. (1975) Comparison of the predicted and observed secondary structure of T4 phage lysozyme. *Biochim. Biophys. Acta (BBA)-Protein Structure*, **405**, 442–451.

54. Gorodkin,J. (2004) Comparing two K-category assignments by a K-category correlation coefficient. *Comput. Biol. Chem.*, **28**, 367–374.

55. Yu,C.-S., Chen,Y.-C., Lu,C.-H. and Hwang,J.-K. (2006) Prediction of protein subcellular localization. *Proteins Struct. Funct. Bioinform.*, **64**, 643–651.

56. Imai,K. and Nakai,K. (2010) Prediction of subcellular locations of proteins: where to proceed?. *Proteomics*, **10**, 3970–3983.

57. Sloma,M.F. and Mathews,D.H. (2016) Exact calculation of loop formation probability identifies folding motifs in RNA secondary structures. *RNA*, **22**, 1808–1818.

58. Tan,Z., Fu,Y., Sharma,G. and Mathews,D.H. (2017) TurboFold II: RNA structural alignment and secondary structure prediction informed by multiple homologs. *Nucleic Acids Res.*, **45**, 11570–11581.

59. Singh,J., Hanson,J., Paliwal,K. and Zhou,Y. (2019) RNA secondary structure prediction using an ensemble of two-dimensional deep neural networks and transfer learning. *Nat. Commun.*, **10**, 5407.

60. Sato,K., Akiyama,M. and Sakakibara,Y. (2021) RNA secondary structure prediction using deep learning with thermodynamic integration. *Nat. Commun.*, **12**, 941.

61. Fu,L., Cao,Y., Wu,J., Peng,Q., Nie,Q. and Xie,X. (2022) UFold: fast and accurate RNA secondary structure prediction with deep learning. *Nucleic Acids Res.*, **50**, e14–e14.

62. Rost,B. (1999) Twilight zone of protein sequence alignments. *Protein Eng.*, **12**, 85–94.

63. Fay,M.P. and Proschan,M.A. (2010) Wilcoxon-Mann-Whitney or t-test? On assumptions for hypothesis tests and multiple interpretations of decision rules. *Statistics surveys*, **4**, 1.

64. Moreno,E., Cloeckaert,A. and Moriyón,I. (2002) Brucella evolution and taxonomy. *Vet. Microbiol.*, **90**, 209–227.

65. Kaack,L.H., Donti,P.L., Strubell,E., Kamiya,G., Creutzig,F. and Rolnick,D. (2022) Aligning artificial intelligence with climate change mitigation. *Nat. Climate Change*, **12**, 518–527.